\documentclass[10pt]{article}
\usepackage{cosmosreport}
\usepackage{flafter}
\usepackage{xspace}

\newcommand{\bench}{ChemArgus\xspace}

\newcommand{\kw}[1]{\textbf{\textcolor{accentdk}{#1}}}

\newcommand{\compA}{full problem\xspace}
\newcommand{\compB}{sequential carry\xspace}
\newcommand{\compC}{oracle scaffolded\xspace}

\begin{document}

\reporttitleblock
%
{\hspace{2mm}\raisebox{-0.27\height}{\includegraphics[height=26pt]{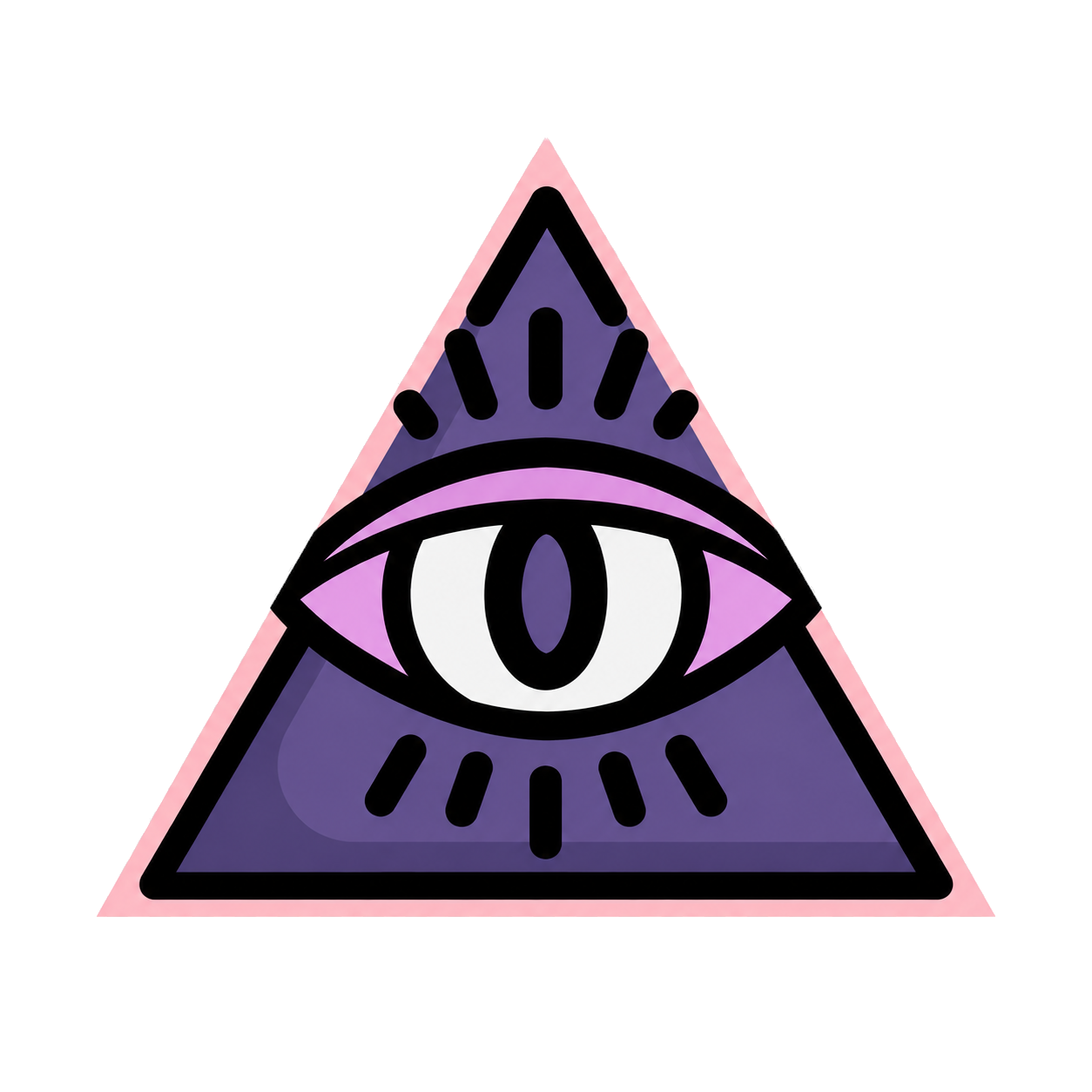}}\hspace{5pt}%
\raisebox{-3.319pt}[0pt][0pt]{\includegraphics[height=15.8794pt]{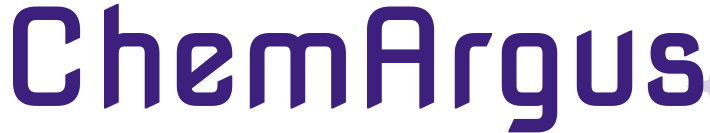}}: Automated Scientific Diagnostics for Lab-Grade Multimodal\\ Chemistry Reasoning: A Frontier Benchmark}
{\scriptsize
 Hanyao Guo$^{\ast}$\quad Mo Cui$^{\ast\dagger}$\quad Zihan Tan$^{\ast}$\quad
 Leixin Sun$^{\ast}$\quad Yichun Wang$^{\dagger}$\quad Yinan Shen$^{\dagger}$\quad Guojun Zhu$^{\dagger}$\quad Jia Jiang\\[1.6pt]
 Erhao Chen\quad Zui Chen\quad Wenzhe Gao\quad Zhihao Yu\quad Qijun Wang\quad Yetian Zhu\quad Jiande Chen\quad Chuhang Pei\\[1.6pt]
 Yao Jiang\quad Yichen Fan\quad Pinhan Wang\quad Xiaorui Pan\quad Qianyi Zhu\quad Yang Liu\quad Mang Ye$^{\ddagger}$\quad Guancheng Wan$^{\ddagger}$\\[2.2pt]
 {\normalfont\scriptsize\color{black!62}$^{\ast}$Equal contribution\quad
  $^{\dagger}$Project leads\quad $^{\ddagger}$Corresponding author}}
{\instblock}
{
%
%
%
Chemistry evaluation has moved past multiple-choice probes toward
open-ended expert problems, yet the benchmarks that define the current
frontier still score answer endpoints alone. Such scoring conflates
memorized facts with stepwise derivation, treats visual and textual
inputs as interchangeable, and confounds long-horizon planning with
stepwise execution. These blind spots prevent any diagnosis of where
the reasoning chain breaks in a model. We present \ab{ChemArgus}, a
competition-level chemistry benchmark drawn from chemistry competition
material, textbooks, and research papers, whose scoring replaces the
answer endpoint with three measurement properties. \ab{Fine-grained
scoring} grades every subquestion point by point with partial
credit, so a score lands at the level of the step rather than the
outcome. \ab{Multimodal capability} is priced on every problem containing
a real image, paired with a controlled text description, so visual
understanding is measured against text on identical tasks. \ab{Diagnostic
attribution} reads paired metrics that share one scoring unit and differ
only in context construction, so a gap between two scores isolates a
single cause, namely integration, propagation, or local reasoning.
GPT-5.6-Sol scores 23.2\% on unassisted whole-problem solving against
28.8\% and 30.4\% under the two sequential metrics, a gap measuring the
integration burden. Through rubric partial credit and these
cause-isolating score gaps, \ab{ChemArgus}\ shows not only whether a model
answers chemistry correctly but where its reasoning chain fails, so that
every score it reports doubles as a diagnosis, a readout of the chain
rather than a verdict, aiming to guide language models step by step on
their path toward chemistry expertise, one corrected step at a time, each step
priced by the rubric, point by point.
}
{\centering
\linkbutton[RoleA]{\faGithub}{ChemArgus}{https://github.com/YunQi572/ChemArgus.git}\hspace{3mm}%
\linkbutton[RoleC]{\faDatabase}{Hugging Face}{https://huggingface.co/datasets/CosmosMind/ChemArgus}\par}

\begin{center}
{\includegraphics[width=\linewidth]{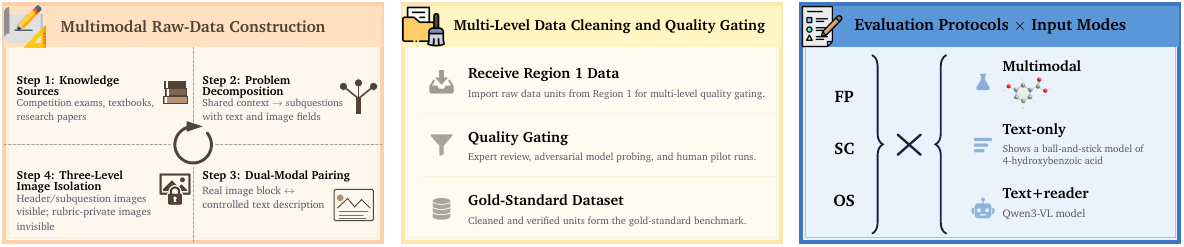}}
\captionof{figure}{\textbf{The pipeline of \bench\ at a glance.} It runs the full flow from problem construction through cleaning and gating to evaluation, a four-step refinement loop with three gating stages for candidate problems, and the protocol structure of every reported run, where \compA (FP) evaluates the whole problem in one pass, \compB (SC) carries the model's own previous answers forward, and \compC (OS) grants the standard intermediate result of each preceding subquestion, one protocol per access policy on the chain.}
\label{fig:pipeline}
\end{center}

\section{Introduction}
\label{sec:intro}


Chemistry is one of the hardest open domains for language models.
Expert-level problems interleave structure reading from images, multi-step
quantitative derivation, and answers a chemist can independently verify,
and current models show real capability~\citep{v3,kimi15} on these
tasks while still trailing expert solvers by large, stable
margins~\citep{chembench,chemeval,chemiq,qcbench,superchem,gpqa,hle,gpt4,
gemini15,o1}, a gap that persists across every suite reported to date. The stronger the models, the more evaluation must
explain what it measures, because improvement can only happen where an
error is precisely located, and a benchmark has to account for its score
rather than merely report it, so the number becomes a diagnosis, one that
names the step where the chain broke and the reason it broke there.

\begin{center}
{\includegraphics[width=\linewidth]{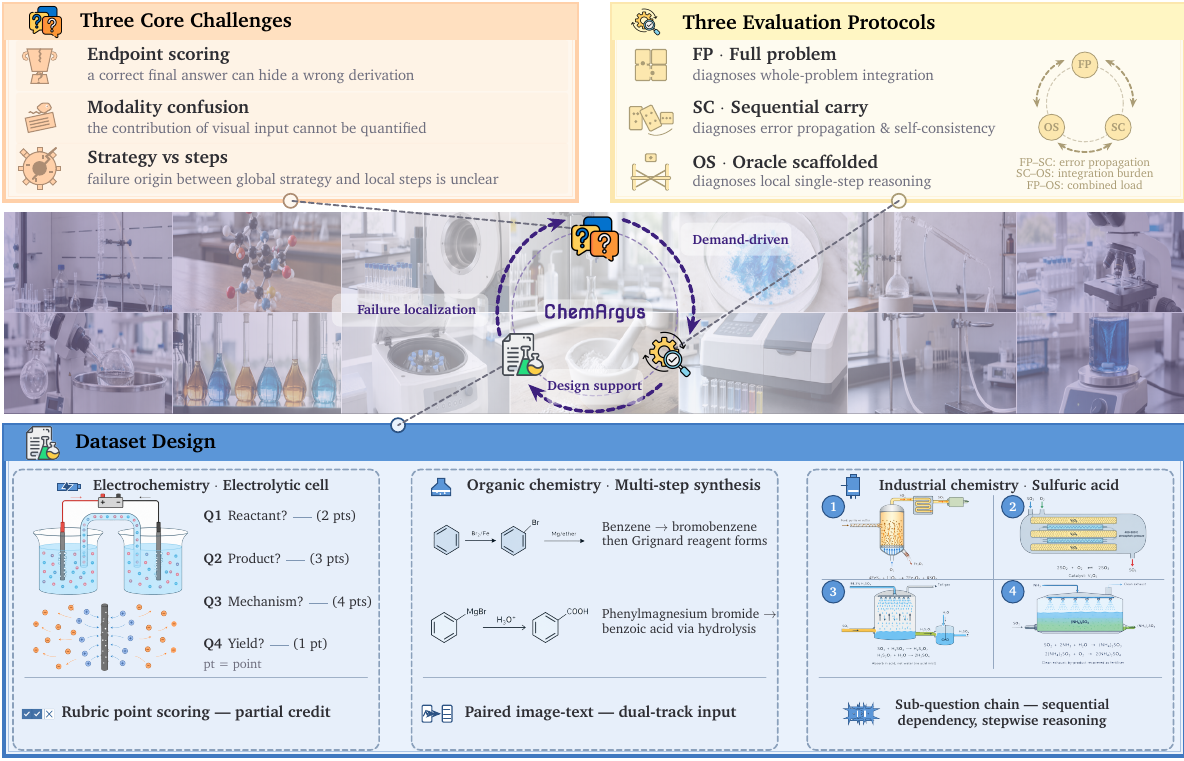}}
\captionof{figure}{\textbf{Overview of \bench.} The complete design at a glance, three measurement challenges on the left, three evaluation protocols sharing one scoring unit on the right, a demand, design, attribution loop across the middle, and three example problem chains from the dataset below, worked end to end within one card, each card showing its graded steps.}
\label{fig:overview}
\end{center}

Expert-level open chemistry problems bundle three demands that existing
benchmarks rarely measure jointly. These demands directly test three core
capabilities required for solving scientific problems. First, models must
\emph{distinguish factual recall from genuine stepwise derivation}. A
correct final answer may be reached through factual recall alone without
genuine stepwise derivation. Endpoint scoring cannot tell
which~\citep{processbench}, the gap \kw{Fine-grained scoring}
closes.
Second, models must \emph{handle visual and semantic inputs in
qualitatively different ways}. Visual structures demand spatial reasoning,
whereas text supplies symbolic logic. However, current evaluations rarely
provide a controlled comparison between modalities and often supply only
text~\citep{chemtable,macbench,mmmu}. Consequently, whether
\kw{Multimodal capability} actually enriches scientific comprehension,
and whether visual input adds understanding beyond text, both remain
unclear. Third, models must \emph{coordinate global planning with local
execution}. Solving a long chain of subquestions requires strategic
decomposition. Scoring only the final outcome conflates a failed
high-level strategy with a single careless step~\citep{scicode}. Whether
failure originates in global strategy or in local execution stays
undecidable by any current score~\citep{scibench}, the confusion
\kw{Diagnostic attribution} resolves. No existing benchmark
measures these three capabilities together with controlled image
visibility and rubric-grounded partial credit, so the field lacks a
\kw{fine-grained}, \kw{multimodal}, \kw{diagnostic} instrument for where
scientific reasoning in models actually breaks, at the step, in the
modality, or in the plan, exactly where the three properties look.

To answer these questions and resolve the identified conflict, we introduce
\bench, a controlled evaluation dataset for high-difficulty chemistry
reasoning, open-ended answering, and multimodal understanding. The
benchmark rests on three measurement properties that the conflict names.
\kw{Fine-grained scoring} decomposes every problem into
subquestions that share one context and grades each subquestion point by
point with partial credit, so one scoring unit spans the whole benchmark
and each score lands at the step rather than the outcome.
\kw{Multimodal capability} is priced through real images entering every
problem together with a controlled text description, so visual
understanding is measured against text on identical tasks.
\kw{Diagnostic attribution} holds that scoring unit fixed while varying
only how the model accesses prior information, so each gap between two
scores names a single cause. The most natural
protocol asks the
model to solve the whole problem in one pass, and \ra{\compA} does exactly
that, measuring integrated solving with nothing granted. Its score,
however, cannot say where a chain broke, because integration and
propagation are confounded in a single pass, so a second protocol,
\rb{\compB}, reveals the subquestions one at a time and carries forward the
answers the model produced earlier, making every propagated error visible.
A third protocol, \rc{\compC}, then grants the standard intermediate result
of each preceding subquestion, so the remaining score measures local
reasoning with propagation removed. Because the three protocols share one
scoring unit, the gaps between their scores are \kw{diagnostic}
quantities, and each gap names a single cause, integration, propagation,
or local reasoning, one gap, one price.


\noindent Our contributions are as follows.
\begin{itemize}[leftmargin=1.9em,itemsep=2pt,topsep=4pt]
\item[\ding{182}] \textbf{\textit{Problem Identification.}} We trace the
interpretability failure of chemistry benchmarks to three named defects,
recall conflated with derivation, modalities treated as interchangeable,
and planning confounded with execution, all properties of the scoring
itself.
\item[\ding{183}] \textbf{\textit{Benchmark and Data.}} We build \bench,
a controlled dataset for high-difficulty chemistry reasoning and
\kw{multimodal} understanding, with rubric partial credit, a controlled
text description for every figure, and rubric-private images kept out of
every model input.
\item[\ding{184}] \textbf{\textit{Evaluation Framework.}} We define three
protocols sharing one scoring unit and evaluate nine models across three
input modes, so each \kw{diagnostic} attribution names a single cause,
integration, propagation, or local reasoning.
\end{itemize}


\section{Related Work}
\label{sec:related}


\subsection{Chemistry Knowledge and Reasoning Benchmarks}
\label{sec:related:chem}

The chemistry benchmarks that established the capability gaps this report
measures share one property. Each scores an answer endpoint, so each can
see that a model is wrong but not where it went wrong. The sequence begins
with broad multiple-choice knowledge probes~\citep{mmlu,arc}, arithmetic
and factual probes~\citep{gsm8k,simpleqa}, multi-task capability
suites~\citep{bigbench,ceval},
their reasoning-focused successor \citep{mmlupro}, eight-task
multiple-choice
chemistry~\citep{chemllmbench}, and \citep{scieval}, roughly 18{,}000
questions across physics, chemistry, and biology with data generated
dynamically against leakage. It reaches its ceiling in expert-written
question sets that remain hard even in closed form, and in
\citep{olympiadbench}, 8{,}476 olympiad-level bilingual multimodal
scientific problems on which the strongest multimodal models of its day
averaged below $18\%$. \citep{gpqa} reports that domain experts reach $65\%$ on its
448 graduate-level questions while the strongest GPT-4 baseline~\citep{gpt4}
manages $39\%$, and \citep{hle} shows frontier models below $10\%$ at
release on
2{,}500 short-answer and multiple-choice questions across 100+ subjects,
chemistry among them. The reasoning-model wave~\citep{o1,r1} and its open-weight
peers~\citep{kimi15,v3} then raised the stakes on process-level
evaluation, and \citep{processbench}
grades each derivation step to identify process errors, the direction
this report's rubrics extend to chemistry. The current generation of
chemistry benchmarks moves to open-ended expert tasks. \citep{chembench}
pairs multiple-choice with open-ended questions across chemistry tasks
and reports GPT-4 at or above estimated human performance on large parts
of the benchmark; \citep{chemeval} structures multi-level chemical
capability evaluation; \citep{chemiq} assesses chemical intelligence over
broad topics; \citep{qcbench} specializes in quantitative problems with
judge-based scoring; and \citep{superchem} adds multimodal reasoning
questions. Each benchmark establishes a real, stable gap between models
and expert chemists. What none of them reports is where a wrong answer
became wrong. Scoring an option or a matched string discards the partial
credit a human examiner would have awarded along the way, so partial
credit stays unlocalized in every one of them, and no score in this
family reports where a wrong answer became wrong, so the gap it measures
stays opaque about the failure point.

\subsection{Multimodal Chemistry Evaluation}
\label{sec:related:multimodal}

Multimodal evaluation prices the visual side of chemistry, and the vision
language model literature supplies the backbones, \citep{llava} opening
the instruction-tuned line that \citep{qwen2vl} scale to any
resolution and \citep{internvl,qwen3vl} carry into interleaved long
contexts. General multimodal benchmarks such as \citep{seedbench,mmbench}
score broad perception and reasoning, \citep{mmmu} and its hardened
successor \citep{mmmupro} include college-level chemistry questions that
cannot be answered without the figure, chart and document suites such as
\citep{chartqa,docvqa} test figure reading directly, visual
mathematics \citep{mathvista,mathverse} shows the same figure dependency
next door, as does \citep{mathvision}. On the chemistry side, \citep{chemtable} measures
recognition and understanding over chemical tables; \citep{chemrag}
benchmarks retrieval-augmented generation in multimodal chemistry
contexts; \citep{macbench} probes multimodal language models on chemistry
and materials tasks and finds systematic failures where the answer
depends on reading a figure; \citep{scienceqa} chains multimodal
reasoning over science questions; and \citep{moleculeqa} checks factual
molecular comprehension across 62{,}000 questions and finds factual
accuracy barely moving even as lexical similarity climbs. Together these
works establish that visual structure reading is genuinely load-bearing.
Models lose points not on general knowledge but on what they see or fail
to see. What none of them provides is a controlled text description of
every figure, so an image-conditioned score cannot be separated from what
the problem text already carries. Image gains are reported, but never
measured against a controlled text counterpart, so modality stays
unpriced across this literature, and no score prices it.

\subsection{Scoring and Protocols}
\label{sec:related:scoring}

Beyond answer scoring and multimodal input, evaluation has also begun to
shape how a solution is graded and how a model is asked. \citep{scibench} grades
college-level science problems against detailed rubrics and finds models
far below undergraduate performance, \citep{qcbench} judges quantitative
chemistry answers, \citep{chembench} judges its open-ended portion with a
model judge, and the judge literature itself matured from pairwise
preference~\citep{llmjudge,lmarena,alpacaeval} through criterion-based
generation~\citep{geval} to systematic judge auditing~\citep{judgebench},
reward calibration~\citep{rewardbench,aurora}, and rubric
alignment~\citep{rubricbench}, while \citep{scicode} shows
that formal answers can be graded deterministically against unit tests
rather than a judge's opinion, and \citep{sciassess} scores scientific
literature analysis against fine-grained rubrics across memorization,
comprehension, and reasoning levels. On the asking side, prompting
studies~\citep{cot} established that how a model is asked changes what it
shows, \citep{scicode} sequences tasks whose steps consume earlier
results, \citep{scibench} studies whether revisiting an earlier mistake
repairs the rest, and agent evaluation moved to real environments,
\citep{agentbench} across interactive settings, \citep{swebench} across
software engineering, \citep{gaia,osworld} across general assistance and
real computer tasks, \citep{webarena} across the open web, and
\citep{toolbench} across tool use, while the standardized harness
\citep{helm} made such evaluation reproducible at scale. On the scientific side,
\citep{scienceagentbench} finds the best agent solving only $32.4\%$ of
its 102 discovery tasks independently, \citep{labbench,labbench2} chains
long laboratory protocols graded by rule matching and simulation
validators, and \citep{discoveryworld} runs discovery agents in a
fictional world that reports progress at checkpoints rather than
per-answer rubric points; \citep{scienceworld} grounds agents in an
interactive science environment, \citep{chemcrow} augments models with
chemistry tools for long task chains, \citep{coscientist} couples a
frontier model with laboratory instruments to plan and execute real
chemistry experiments. These efforts establish that rubric-level scoring,
deterministic checks, and multi-step evaluation are each feasible at
scale. What they leave open is the control surface. No benchmark splits a
rubric into deterministic and judge-scored points, isolates judge-facing
media from the model, or varies the protocol alone while holding the
scoring unit fixed, so the graded artifact stays uncontrolled and
headline numbers stay protocol-dependent.

Two gaps persist across the work reviewed above. First, no existing chemistry
benchmark combines rubric-localized partial credit, a controlled modality
contrast, and protocol separation on a single scoring unit, so none can
localize where a reasoning chain fails. Second, most chemistry benchmarks
remain text-only, and the few that accept images treat modality as an
add-on rather than a measured part of the design, while access control
against training-data contamination is handled as an
stays, the exceptions such as
\citep{labbench} private share under canary monitoring and
\citep{discoveryworld} fictional setting being dataset defenses rather
than measurement design, so the control surface stays open. \bench\ is
multimodal throughout, with real image input on every problem, and
access control built into the measurement design. \bench\ targets exactly
these two gaps.

\section{Benchmark Design}
\label{sec:benchmark}


\noindent\textbf{Overview.} \bench\ is built so that a score is not a single
number a solution earns but a position along the chain a solution travels.
Every subquestion is graded against a rubric with partial credit, the visual
content a model sees is isolated from the visual content used to grade, and
the same scoring unit is instantiated under three protocols that differ only
in how the model's context is assembled. This section develops the four
design layers in order. \Cref{sec:benchmark:task} fixes the data protocol
that normalizes every problem into one JSON form, including which images are
model-visible and which are rubric-private; \Cref{sec:benchmark:qc} describes
the construction and quality-control pipeline every problem passes; \Cref{sec:benchmark:protocols} defines the three evaluation protocols
\ra{\compA}, \rb{\compB}, \rc{\compC} and states precisely what a gap between
two of them isolates; and \Cref{sec:benchmark:scoring} specifies the
dual-track grading pipeline, deterministic chemistry-equivalence checks for
formal answers and an audited independent judge for flexible ones, so the
two answer kinds never share a grader. \Cref{fig:overview} draws the whole
design at a glance, the three challenges, the three protocols, and the
example problem chains joined by one demand, design, attribution loop
that closes on itself, each score feeding the next diagnosis.

\begin{table}[t]
\centering
\caption{Comparison of \bench against prior chemistry and adjacent scientific benchmarks on five properties. ``image input'' reports images as real multimodal blocks, and ``image-text contrast'' reports every figure paired with controlled text. ``rubric point partial credit'' reports partial credit per rubric point, and ``subquestion chains'' report problems split into shared-context subquestions. Finally, ``protocol separation'' reports one scoring unit under several protocols. \bench\ is the only benchmark providing all five, green checkmarks for presence and red crosses for absence.}
\label{tab:benchmark_comparison}
\small

\renewcommand{\arraystretch}{1.25}
\begin{tabularx}{\textwidth}{>{\raggedright\arraybackslash\hsize=2.0\hsize}X >{\centering\arraybackslash\hsize=0.8\hsize}X >{\centering\arraybackslash\hsize=0.8\hsize}X >{\centering\arraybackslash\hsize=0.8\hsize}X >{\centering\arraybackslash\hsize=0.8\hsize}X >{\centering\arraybackslash\hsize=0.8\hsize}X}
\toprule
\headrow \makecell[c]{\bfseries Benchmark} & \makecell[c]{\bfseries Image\\input} & \makecell[c]{\bfseries Image-text\\contrast} & \makecell[c]{\bfseries Rubric\\partial credit} & \makecell[c]{\bfseries Subquestion\\chains} & \makecell[c]{\bfseries Protocol\\separation} \\
\midrule
SciBench \citep{scibench} \textcolor{gray}{[ICML24]} & \textcolor{posgreen}{\ding{51}} & \textcolor{negred}{\ding{55}} & \textcolor{negred}{\ding{55}} & \textcolor{negred}{\ding{55}} & \textcolor{negred}{\ding{55}} \\
\bandrow SciCode \citep{scicode} \textcolor{gray}{[NeurIPS24]} & \textcolor{negred}{\ding{55}} & \textcolor{negred}{\ding{55}} & \textcolor{negred}{\ding{55}} & \textcolor{posgreen}{\ding{51}} & \textcolor{negred}{\ding{55}} \\
GPQA \citep{gpqa} \textcolor{gray}{[arXiv23]} & \textcolor{negred}{\ding{55}} & \textcolor{negred}{\ding{55}} & \textcolor{negred}{\ding{55}} & \textcolor{negred}{\ding{55}} & \textcolor{negred}{\ding{55}} \\
\bandrow DiscoveryWorld \citep{discoveryworld} \textcolor{gray}{[NeurIPS24]} & \textcolor{negred}{\ding{55}} & \textcolor{negred}{\ding{55}} & \textcolor{negred}{\ding{55}} & \textcolor{negred}{\ding{55}} & \textcolor{negred}{\ding{55}} \\
SciEval \citep{scieval} \textcolor{gray}{[AAAI24]} & \textcolor{negred}{\ding{55}} & \textcolor{negred}{\ding{55}} & \textcolor{negred}{\ding{55}} & \textcolor{negred}{\ding{55}} & \textcolor{negred}{\ding{55}} \\
\bandrow ChemBench \citep{chembench} \textcolor{gray}{[arXiv24]} & \textcolor{negred}{\ding{55}} & \textcolor{negred}{\ding{55}} & \textcolor{negred}{\ding{55}} & \textcolor{negred}{\ding{55}} & \textcolor{negred}{\ding{55}} \\
ChemLLMBench \citep{chemllmbench} \textcolor{gray}{[NeurIPS23]} & \textcolor{posgreen}{\ding{51}} & \textcolor{negred}{\ding{55}} & \textcolor{negred}{\ding{55}} & \textcolor{negred}{\ding{55}} & \textcolor{negred}{\ding{55}} \\
\bandrow ChemEval \citep{chemeval} \textcolor{gray}{[ICLR26]} & \textcolor{posgreen}{\ding{51}} & \textcolor{negred}{\ding{55}} & \textcolor{negred}{\ding{55}} & \textcolor{negred}{\ding{55}} & \textcolor{negred}{\ding{55}} \\
ChemIQ \citep{chemiq} \textcolor{gray}{[JCIM26]} & \textcolor{negred}{\ding{55}} & \textcolor{negred}{\ding{55}} & \textcolor{negred}{\ding{55}} & \textcolor{negred}{\ding{55}} & \textcolor{negred}{\ding{55}} \\
\bandrow SUPERChem \citep{superchem} \textcolor{gray}{[arXiv25]} & \textcolor{posgreen}{\ding{51}} & \textcolor{posgreen}{\ding{51}} & \textcolor{negred}{\ding{55}} & \textcolor{negred}{\ding{55}} & \textcolor{negred}{\ding{55}} \\
QCBench \citep{qcbench} \textcolor{gray}{[JCIM25]} & \textcolor{negred}{\ding{55}} & \textcolor{negred}{\ding{55}} & \textcolor{negred}{\ding{55}} & \textcolor{negred}{\ding{55}} & \textcolor{negred}{\ding{55}} \\
\bandrow MaCBench \citep{macbench} \textcolor{gray}{[NatCompSci25]} & \textcolor{posgreen}{\ding{51}} & \textcolor{posgreen}{\ding{51}} & \textcolor{negred}{\ding{55}} & \textcolor{negred}{\ding{55}} & \textcolor{negred}{\ding{55}} \\
ChemTable \citep{chemtable} \textcolor{gray}{[KDD26]} & \textcolor{posgreen}{\ding{51}} & \textcolor{negred}{\ding{55}} & \textcolor{negred}{\ding{55}} & \textcolor{negred}{\ding{55}} & \textcolor{negred}{\ding{55}} \\
\bandrow ChemRAG-Bench \citep{chemrag} \textcolor{gray}{[COLM25]} & \textcolor{negred}{\ding{55}} & \textcolor{negred}{\ding{55}} & \textcolor{negred}{\ding{55}} & \textcolor{negred}{\ding{55}} & \textcolor{negred}{\ding{55}} \\
\midrule
\textbf{\bench} & \textbf{\textcolor{posgreen}{\ding{51}}} & \textbf{\textcolor{posgreen}{\ding{51}}} & \textbf{\textcolor{posgreen}{\ding{51}}} & \textbf{\textcolor{posgreen}{\ding{51}}} & \textbf{\textcolor{posgreen}{\ding{51}}} \\
\bottomrule
\end{tabularx}
\end{table}

\noindent Table~\ref{tab:benchmark_comparison} compares \bench against the
benchmarks reviewed in \Cref{sec:related} on five properties, real image
input, controlled image-text contrast, rubric point partial credit,
subquestion chains, and protocol separation. On the boundary cases,
ChemBench~\citep{chembench} contains no images and represents molecules as
SMILES; ChemEval~\citep{chemeval} offers text and multimodal versions that
are task-level rather than per-figure descriptions;
SUPERChem~\citep{superchem} is multiple choice with dual-format questions;
MaCBench~\citep{macbench} is mostly VQA and multiple choice;
DiscoveryWorld~\citep{discoveryworld} offers an optional 2D visual
interface that is not a data figure, and its reports carry no rubric
points.


\subsection{Task Formulation and Data Protocol}
\label{sec:benchmark:task}

Competition chemistry problems do not fit answer-level scoring because they
do not have answer-level solutions. A single problem is a \emph{chain} of
subquestions that share a context, reuse earlier results, and are graded
point by point. The data protocol therefore preserves that chain. Every
problem is normalized into one JSON unit consisting of a shared
\texttt{header} that gives the background and the substances involved,
with its header images, followed by an ordered list of
\texttt{subquestions}. Each subquestion carries its open-ended text,
model-visible subquestion images, a positive integer full mark,
its \texttt{rubric} as an ordered list of scoring points, and
rubric-private rubric images; stable problem and subquestion IDs
identify every unit, and knowledge \texttt{tags}
link each subquestion to a fixed category vocabulary. No extra fields are
permitted, so two instantiations of the benchmark can only differ in
content, never in structure, and their results remain directly comparable
across instantiations and across runs.

Two properties of the protocol are load-bearing for what this report
measures. First, \emph{images are data, not decoration}. Every image is
delivered to the model as a real multimodal input block, never as a path or
a placeholder string, and every image also carries a curated text
description of what a competent reader would extract. The description
supports the text-only protocol in \Cref{sec:benchmark:protocols} and makes
the modality contrast of \Cref{sec:exp} a controlled one. The same content
reaches the model twice, once as pixels and once as prose. Second,
\emph{image visibility is a design class, not a leak}. The three image
fields are split into model-visible media, header images and subquestion
images, and rubric-private media, rubric images, scoring structures,
reference answers, and boundary diagrams.
Rubric-private
images never enter any model input under any protocol, so a model cannot
read its grading key from the problem itself, under any protocol or input
mode, in any reported run.

\subsection{Dataset Construction}
\label{sec:benchmark:qc}

\bench\ is built as a measurement instrument for expert-level chemistry
reasoning rather than a repository of hard questions. It evaluates
whether a model can form a consistent and verifiable reasoning chain
across molecular structure, reaction mechanism, symbolic representation,
and expert scoring logic throughout the solution path. We construct every
problem from first principles, drawing on chemistry competition
knowledge, textbooks, and research literature. Each problem must satisfy
three conditions, namely \emph{chemical correctness}, \emph{derivability
from the statement}, and \emph{sufficient discrimination between factual
recall and genuine reasoning}.

Candidate problems are filtered through expert review, adversarial model
probing, and student pilots. Only those that satisfy correctness,
scorability, and meaningful challenge enter the final set.
\Cref{fig:pipeline} draws the construction and evaluation pipeline, the
four-step refinement loop, the three gating stages, and the protocol
structure every reported run passes through. The final dataset is
delivered as structured JSON units containing problem identifiers,
question text, model-visible images with text descriptions, rubrics, and
rubric-private images. This structure enables fine-grained diagnosis of
where the reasoning chain breaks in a model, down to the subquestion and
the protocol that cost the score, exactly the diagnosis promised.

\subsection{Three Evaluation Protocols}
\label{sec:benchmark:protocols}

The three protocols are the instrument this benchmark is built around. They
ask the same subquestions and are graded by the same scoring unit; they
differ only in what the model has in its context when it answers each
subquestion, which is what lets a score gap between two protocols name a
cause instead of confound one.

\noindent\textbf{\textcolor{accent}{Full problem}}. The model receives the complete problem, shared
header, images, and all subquestions at once, and answers everything in one
pass. This is the unassisted whole-problem setting. Global planning,
cross-subquestion integration, and long-context organization are fully
loaded, and nothing else is.

\noindent\textbf{\textcolor{accent}{Sequential carry}}. The model answers subquestions in order inside one
session; each subquestion sees the model's own previous answers but never
the standard results or the rubric. This is the natural sequential solving
setting, and it is where self-consistency and error propagation become
visible. A wrong structure in an early subquestion is what every later
subquestion inherits and must then live with.

\noindent\textbf{\textcolor{accent}{Oracle scaffolded}}. The model answers each subquestion in isolation,
receiving the standard intermediate result of the immediately preceding
subquestion as a scaffold. Upstream correctness is therefore granted rather
than earned, and the remaining score measures local capability on the
subquestion itself.

Because the scoring unit is identical across protocols, the three gaps are
diagnostic quantities. FP versus SC prices \emph{sequential error
propagation}, what a model loses by carrying its own mistakes forward. SC
versus OS prices \emph{integration burden} under propagation, how much the
standard intermediate results rescue. FP versus OS prices the combined
\emph{integration and propagation} load. A benchmark that reports all three
scores has, in effect, localized the failure of a solution chain before any
qualitative analysis begins.

\subsection{Rubric-Grounded Scoring}
\label{sec:benchmark:scoring}

Grading answers the chain, not its end. Each subquestion's rubric decomposes
it into executable scoring points, and a model earns each point
independently, so a solution right in nine of ten derivations scores nine of
ten points' worth rather than zero. The grading pipeline is deliberately
dual-track, because chemistry answers come in two kinds and each kind
deserves the strongest available verifier~\citep{scibench,qcbench}.

\emph{Formal answers}, structures, SMILES, equations, and computed values,
pass deterministic normalization and chemistry-equivalence checks~\citep{scicode}
before any judge sees them. Two equivalent SMILES are equivalent in the grade, and a
structure is checked as a chemical object rather than as a string match.
\emph{Flexible answers}, mechanistic explanations and experimental
reasoning, are graded by an independent judge session that evaluates each
rubric point
against the answer, returning a per-point verdict with evidence and a
confidence level~\citep{scibench,llmjudge,geval,rubricbench}.
Low-confidence verdicts and boundary samples are routed to human review. Every grading interaction, problem
input, model output, judge request, verdict, and summary is archived, so a
published score can be traced to the exact rubric points that earned it.
Together the two tracks keep what judges are good at, open-ended chemical
judgment, and take away what they are bad at, structure equivalence, and
the archive turns the rubric from a scoring device into an audit device
that traces every point to its evidence.


\section{Evaluation}
\label{sec:setup}


\subsection{Models}
\label{sec:eval:models}

We evaluate nine frontier models, numbered (1) through (9) as follows.
(1) GPT-5.6-Sol from OpenAI, a flagship reasoning model with 1M token
context designed for scientific and agentic workloads;
(2) Gemini 3.8 Flash from Google, featuring 1M token context and native
multimodal reasoning; (3) Claude Opus 5 from Anthropic,
capable of extended agentic operation on hard knowledge
work; (4) Qwen3.8-Max from Alibaba, with
2.4T parameters and 1M token context optimized for coding and
research; (5) Kimi K3 from Moonshot, with 2.8T parameters,
1M token context, and native vision understanding; (6)
Grok 4.6 from xAI, with 500K context focused on long-running agentic
tasks and visual reasoning; (7) MiniMax M3 from MiniMax, a native
multimodal open-weight model with 1M token context;
(8) DeepSeek V4 Pro from DeepSeek, a 1.6T MoE model with 1M token context
delivering top-tier open-source performance; and (9)
GLM-5.2 from Zhipu AI, an open-source model with 1M token context strong
on long-horizon tasks. All models run at default settings
with no task-specific fine-tuning, so the scores reflect off-the-shelf
models.

\subsection{Scoring}
\label{sec:eval:scoring}

All models are evaluated within the open-source Pi agent harness, except
Claude Opus 5, which runs in the Claude Code harness, and both provide a
uniform calling interface and archive per-run image and tool audits. All runs use zero prompting, delivering only the problem content
required by the protocol, so no score rests on prompting
skill~\citep{cot}. No few-shot examples, reasoning instructions, tools, or
web search are provided, so every score measures the model's own
chemistry alone.

All scores are normalized rubric scores, the share of rubric points
earned. Grading is performed per subquestion by an independent LLM judge,
GPT-5.6-Sol, under the audited dual-track pipeline described in
\Cref{sec:benchmark:scoring}. Rubric-based LLM judging is standard
practice by now~\citep{llmjudge,geval,judgebench,rubricbench}, and the
judge operates strictly against the rubric points, never against a
holistic impression. Formal answers pass deterministic
chemistry-equivalence checks before the judge session, while flexible
answers are evaluated by the judge against each rubric point. Human
spot-checking validates judge verdicts, and the agreement between judge
and human scores is high, strong enough to report the judge numbers
alone, without a human pass over every answer.

\subsection{Input Modes}
\label{sec:eval:protocols}

The benchmark adopts three input modes. Multimodal runs receive every
model-visible image as a real image block. Text-only runs receive the text
description of each image and no image blocks. Text+reader runs pair a
text model with the Qwen3-VL image reader~\citep{qwen3vl}, where the model
autonomously decides when and how to invoke the reader for image
recognition, converting visual content into text before processing.

On top of these input modes, the benchmark adopts the three metrics of
\Cref{sec:benchmark:protocols}. \ra{\compA} presents the entire problem at
once, \rb{\compB} reveals subquestions in order and carries forward the
model's own previous answers, and \rc{\compC} injects the standard
intermediate result of each preceding subquestion. Because the three
metrics share one scoring unit, a score gap between two of them isolates
a single cause, the integration burden, the error propagation, or the
local reasoning, one cause per gap.

The three input modes, combined with the three metrics, directly price
the three defects identified in \Cref{sec:intro}. The rubric scores every
subquestion point by point, so factual recall cannot masquerade as genuine
derivation, addressing the recall defect. Comparing multimodal against
text-only runs quantifies the contribution of real image input,
addressing the modality defect. The three metrics separate integration
burden, error propagation, and local reasoning, addressing the protocol
defect. Text+reader runs add an additional contrast, measuring what the
reader module contributes beyond text descriptions. Results for all
configurations appear in \Cref{sec:exp}.

\section{Results}
\label{sec:exp}


\subsection{Overall Results}
\label{sec:exp:overall}

The roster now reports twenty-five completed runs across nine models and
three input modes, and the census is complete. Each input mode prices a
different competence, so the same model
appears three times in three roles, the multimodal row prices its own
vision, the text row prices its reading of curated descriptions, and the
reader row prices its delegation to an external vision module, and the
three rows of one model are three different abilities rather than three
snapshots of one.

\begin{figure}[htb]
\centering
\includegraphics[width=\linewidth]{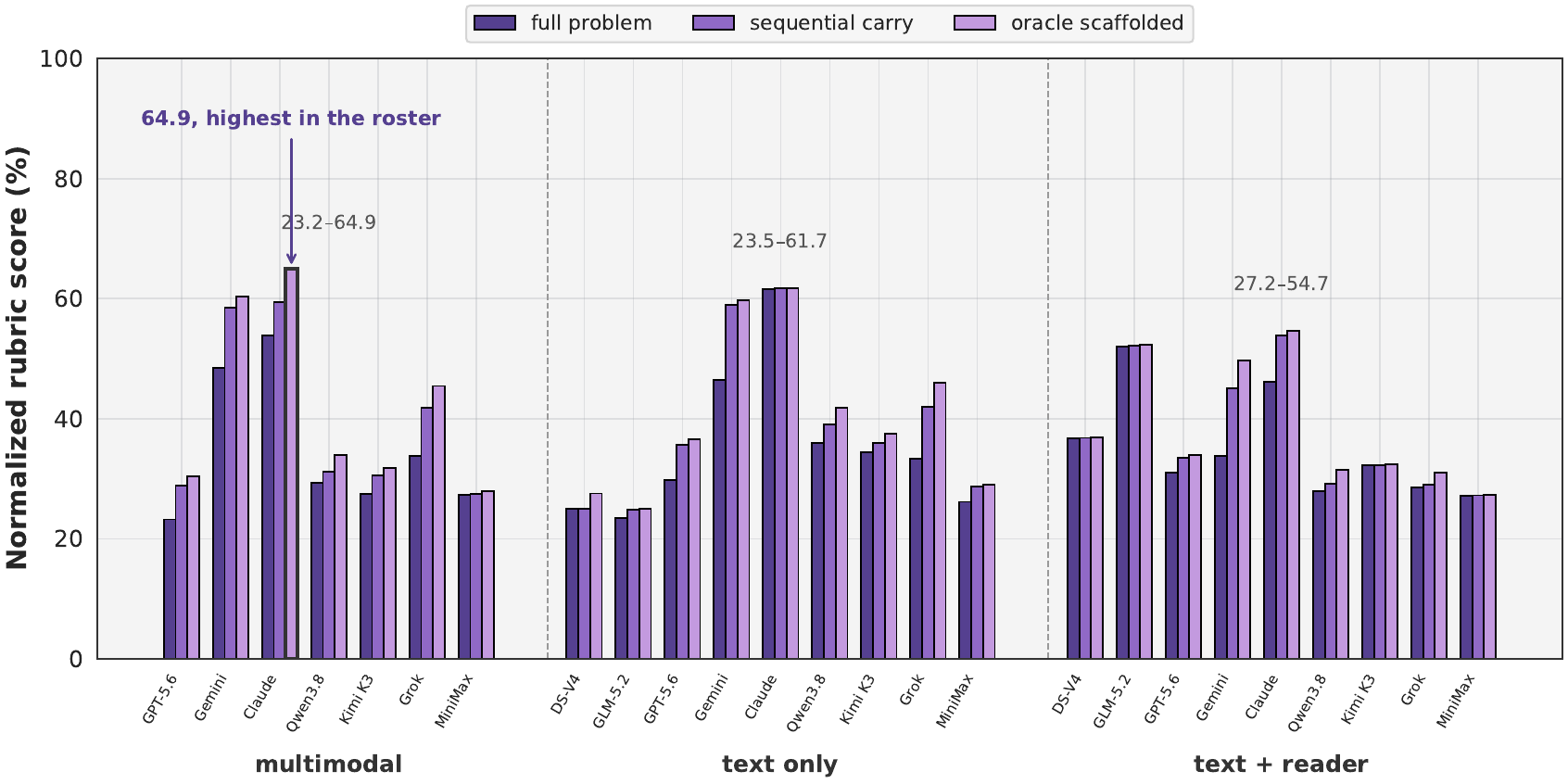}
\caption{Every reported run under the three metrics, grouped by input mode, normalized rubric score in \%.}
\label{fig:roster_profile}
\end{figure}

The reported runs span a wide band, and Table~\ref{tab:roster_results}
with \Cref{fig:roster_profile} give the full picture. Unassisted
whole-problem solving is the lowest metric in every one of the
twenty-five runs, and the ladder climbs from the whole problem to the step
in all of them, so the integration burden measured by
\Cref{sec:benchmark:protocols} is paid by every model under every input
mode. The profile picture reads the roster at a glance, the multimodal
panel spans 23.2 to 64.9, the text panel 23.5 to 61.7, and the reader
panel 27.2 to 54.7. Claude Opus 5 leads every metric, 53.8 on the whole
problem and 64.9 under oracle scaffolded from multimodal input, ahead of
every other run, and Gemini 3.8 Flash follows from real images,
48.5 on the whole problem and 60.3 under oracle scaffolded, and its text
row trails its image row on two of
the three metrics, the first model in the roster for which reading the
images beats reading the descriptions. GLM-5.2 still leads the reader
panel on the whole problem at
52.0, and the leaders move with the input mode and with the metric, so
the benchmark produces one ranking per condition rather than one winner.

\begin{table}[t]
\centering
\small
\caption{Results on the expanded model roster, normalized rubric score in \%, rounded to one decimal. Bold marks the best score per metric. Multimodal runs receive real image blocks, text only runs receive curated text descriptions with zero image blocks, and text + reader runs pair a text model with the Qwen3-VL image reader, which converts each image to text.}
\label{tab:roster_results}
\begin{tabularx}{\textwidth}{>{\raggedright\arraybackslash\hsize=1.0\hsize}X >{\centering\arraybackslash\hsize=0.7\hsize}X >{\centering\arraybackslash\hsize=1.1\hsize}X >{\centering\arraybackslash\hsize=1.1\hsize}X >{\centering\arraybackslash\hsize=1.1\hsize}X}
\toprule
\headrow \multicolumn{1}{c}{Model} & Input & \ra{\compA} & \rb{\compB} & \rc{\compC} \\
\midrule
GPT-5.6-Sol & multimodal & 23.2 & 28.8 & 30.4 \\
\bandrow Gemini 3.8 Flash & multimodal & 48.5 & 58.5 & 60.3 \\
Claude Opus 5 & multimodal & 53.8 & 59.4 & \textbf{64.9} \\
\bandrow Qwen3.8-Max & multimodal & 29.3 & 31.2 & 34.0 \\
Kimi K3 & multimodal & 27.5 & 30.6 & 31.8 \\
\bandrow Grok 4.6 & multimodal & 33.8 & 41.8 & 45.5 \\
MiniMax M3 & multimodal & 27.3 & 27.5 & 27.9 \\
\midrule
DeepSeek V4 Pro & text only & 25.0 & 25.0 & 27.6 \\
\bandrow GLM-5.2 & text only & 23.5 & 24.9 & 25.0 \\
GPT-5.6-Sol & text only & 29.8 & 35.6 & 36.5 \\
\bandrow Gemini 3.8 Flash & text only & 46.5 & 59.0 & 59.7 \\
Claude Opus 5 & text only & \textbf{61.5} & \textbf{61.7} & 61.7 \\
\bandrow Qwen3.8-Max & text only & 36.0 & 39.0 & 41.9 \\
Kimi K3 & text only & 34.5 & 36.0 & 37.5 \\
\bandrow Grok 4.6 & text only & 33.4 & 42.0 & 46.0 \\
MiniMax M3 & text only & 26.2 & 28.7 & 29.1 \\
\midrule
DeepSeek V4 Pro & text + reader & 36.7 & 36.8 & 37.0 \\
\bandrow GLM-5.2 & text + reader & 52.0 & 52.2 & 52.4 \\
GPT-5.6-Sol & text + reader & 31.1 & 33.6 & 34.0 \\
\bandrow Gemini 3.8 Flash & text + reader & 33.8 & 45.1 & 49.7 \\
Claude Opus 5 & text + reader & 46.2 & 53.9 & 54.7 \\
\bandrow Qwen3.8-Max & text + reader & 27.9 & 29.2 & 31.5 \\
Kimi K3 & text + reader & 32.3 & 32.3 & 32.4 \\
\bandrow Grok 4.6 & text + reader & 28.6 & 29.0 & 31.1 \\
MiniMax M3 & text + reader & 27.2 & 27.2 & 27.3 \\
\bottomrule
\end{tabularx}
\vspace{2pt}
\end{table}

The ladder is universal, but its height varies by family. Gemini 3.8
Flash and Grok 4.6 pay the steepest ones, 13.2 and 12.6 percentage
points (pp) from text alone, and Claude Opus 5 pays 11.1 pp with images,
while GLM-5.2 climbs
only 1.5 pp, Claude Opus 5 climbs 0.2 pp from text,
and the flattened reader runs climb under 0.6 pp each, so the integration
burden is a per-model price rather than a constant tax. The reader
subsection returns to the one exception, Gemini 3.8 Flash under the
reader climbs 15.9 pp, the steepest ladder in the whole roster, sharper
than any of its own text runs.

\begin{figure}[htb]
\centering
\includegraphics[width=\linewidth]{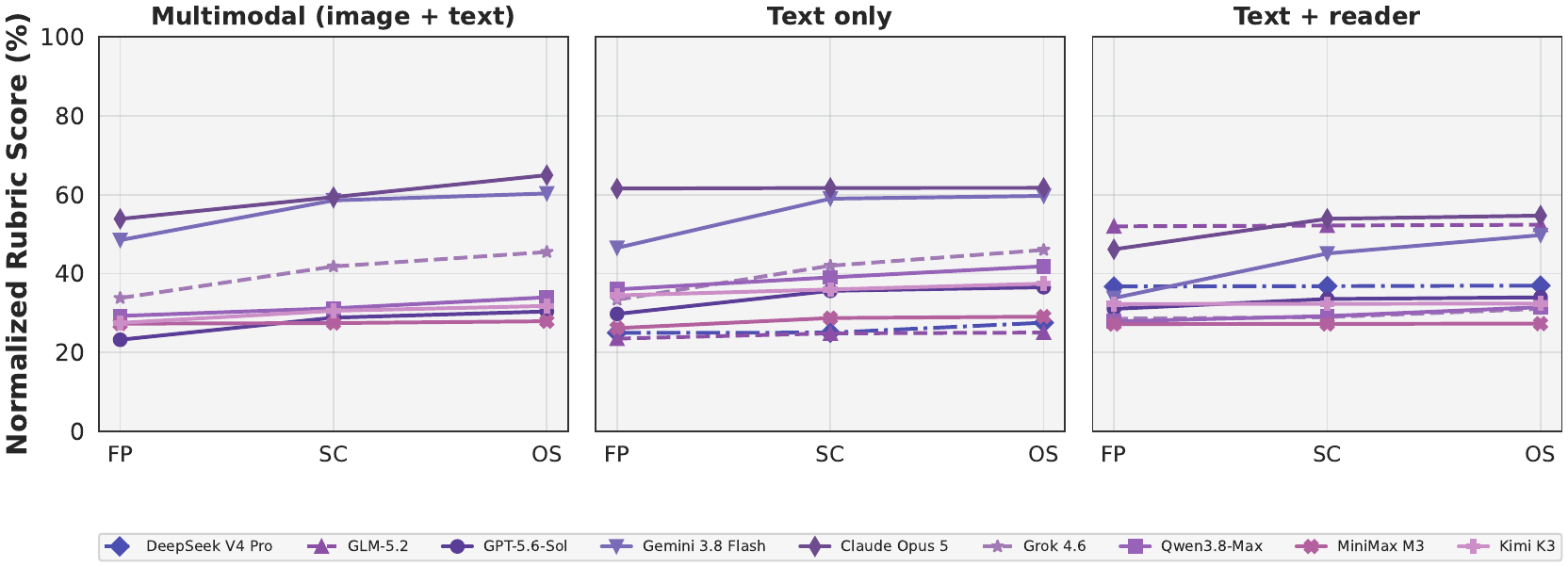}
\caption{Score trajectory across the three metrics for every reported run, one panel per input mode, with scores as normalized rubric percentages.}
\label{fig:protocol_trajectory}
\end{figure}

\Cref{fig:protocol_trajectory} shows where each family bends. The
multimodal runs split into bands, the Claude Opus 5 line riding the top,
Gemini 3.8 Flash and Grok 4.6 the band below it, and the other four
climbing in parallel at the bottom,
the text runs fan out from a flat pair at the bottom to the Gemini ridge,
with the Claude Opus 5 text line lying flat above it,
and the reader runs split in two, GLM-5.2 riding a plateau
above fifty, the Claude Opus 5 reader line climbing from the forties into
the fifties, the Gemini reader line climbing into the forties, and the
rest staying in the twenties and thirties.

\subsection{Modality Contrast}
\label{sec:exp:modality}

The modality contrast now splits the roster in two. For Gemini 3.8
Flash, Grok 4.6, and MiniMax M3, the whole problem favors real images, by
2.0, 0.4, and 1.1 pp, while GPT-5.6-Sol, Qwen3.8-Max, Kimi K3, and
Claude Opus 5
lose 6.5 to 7.7 pp on the whole problem by reading them, as
Table~\ref{tab:modality_gap} shows, so the modality premium is a
property of the model's vision, paid by the weak readers and earned by
the strong ones.

\begin{table}[t]
\centering
\small
\caption{Modality contrast for the seven models reported under both inputs, normalized rubric score in \%, where mm reports the multimodal run with real image blocks and text reports the text only run with curated text descriptions. $\Delta$ reports text only minus multimodal, in percentage points, positive values favor text only. Bold marks the best input score per metric.}
\label{tab:modality_gap}
\begin{tabularx}{\textwidth}{>{\raggedright\arraybackslash\hsize=1.6\hsize}X >{\centering\arraybackslash\hsize=0.93\hsize}X >{\centering\arraybackslash\hsize=0.93\hsize}X >{\centering\arraybackslash\hsize=0.93\hsize}X >{\centering\arraybackslash\hsize=0.93\hsize}X >{\centering\arraybackslash\hsize=0.93\hsize}X >{\centering\arraybackslash\hsize=0.93\hsize}X >{\centering\arraybackslash\hsize=0.93\hsize}X >{\centering\arraybackslash\hsize=0.93\hsize}X >{\centering\arraybackslash\hsize=0.93\hsize}X}
\toprule
\headrow Model & \multicolumn{3}{c}{\ra{\compA}} & \multicolumn{3}{c}{\rb{\compB}} & \multicolumn{3}{c}{\rc{\compC}} \\
\cmidrule(lr){2-4}\cmidrule(lr){5-7}\cmidrule(lr){8-10}
\headrow & mm & text & $\Delta$ & mm & text & $\Delta$ & mm & text & $\Delta$ \\
\midrule
GPT-5.6-Sol & 23.2 & \textbf{29.8} & \textcolor{posgreen}{+6.5} & 28.8 & \textbf{35.6} & \textcolor{posgreen}{+6.8} & 30.4 & \textbf{36.5} & \textcolor{posgreen}{+6.1} \\
\bandrow Gemini 3.8 Flash & \textbf{48.5} & 46.5 & \textcolor{negred}{-2.0} & 58.5 & \textbf{59.0} & \textcolor{posgreen}{+0.4} & \textbf{60.3} & 59.7 & \textcolor{negred}{-0.6} \\
Claude Opus 5 & 53.8 & \textbf{61.5} & \textcolor{posgreen}{+7.7} & 59.4 & \textbf{61.7} & \textcolor{posgreen}{+2.3} & \textbf{64.9} & 61.7 & \textcolor{negred}{-3.2} \\
\bandrow Qwen3.8-Max & 29.3 & \textbf{36.0} & \textcolor{posgreen}{+6.7} & 31.2 & \textbf{39.0} & \textcolor{posgreen}{+7.8} & 34.0 & \textbf{41.9} & \textcolor{posgreen}{+7.9} \\
Kimi K3 & 27.5 & \textbf{34.5} & \textcolor{posgreen}{+7.0} & 30.6 & \textbf{36.0} & \textcolor{posgreen}{+5.4} & 31.8 & \textbf{37.5} & \textcolor{posgreen}{+5.7} \\
\bandrow Grok 4.6 & \textbf{33.8} & 33.4 & \textcolor{negred}{-0.4} & 41.8 & \textbf{42.0} & \textcolor{posgreen}{+0.2} & 45.5 & \textbf{46.0} & \textcolor{posgreen}{+0.5} \\
MiniMax M3 & \textbf{27.3} & 26.2 & \textcolor{negred}{-1.1} & 27.5 & \textbf{28.7} & \textcolor{posgreen}{+1.3} & 27.9 & \textbf{29.1} & \textcolor{posgreen}{+1.2} \\
\bottomrule
\end{tabularx}
\vspace{2pt}
\end{table}

The table compresses each model into nine cells, and the figure keeps the
two input scores on one span inside every metric, so the deltas read as
distances rather than as arithmetic. The controlled text
description matters here, it is written once per figure by the benchmark
authors rather than generated by the model, so both inputs describe the
same content from the same source, and the gap measures the modality
alone.

\begin{figure}[htb]
\centering
\includegraphics[width=\linewidth]{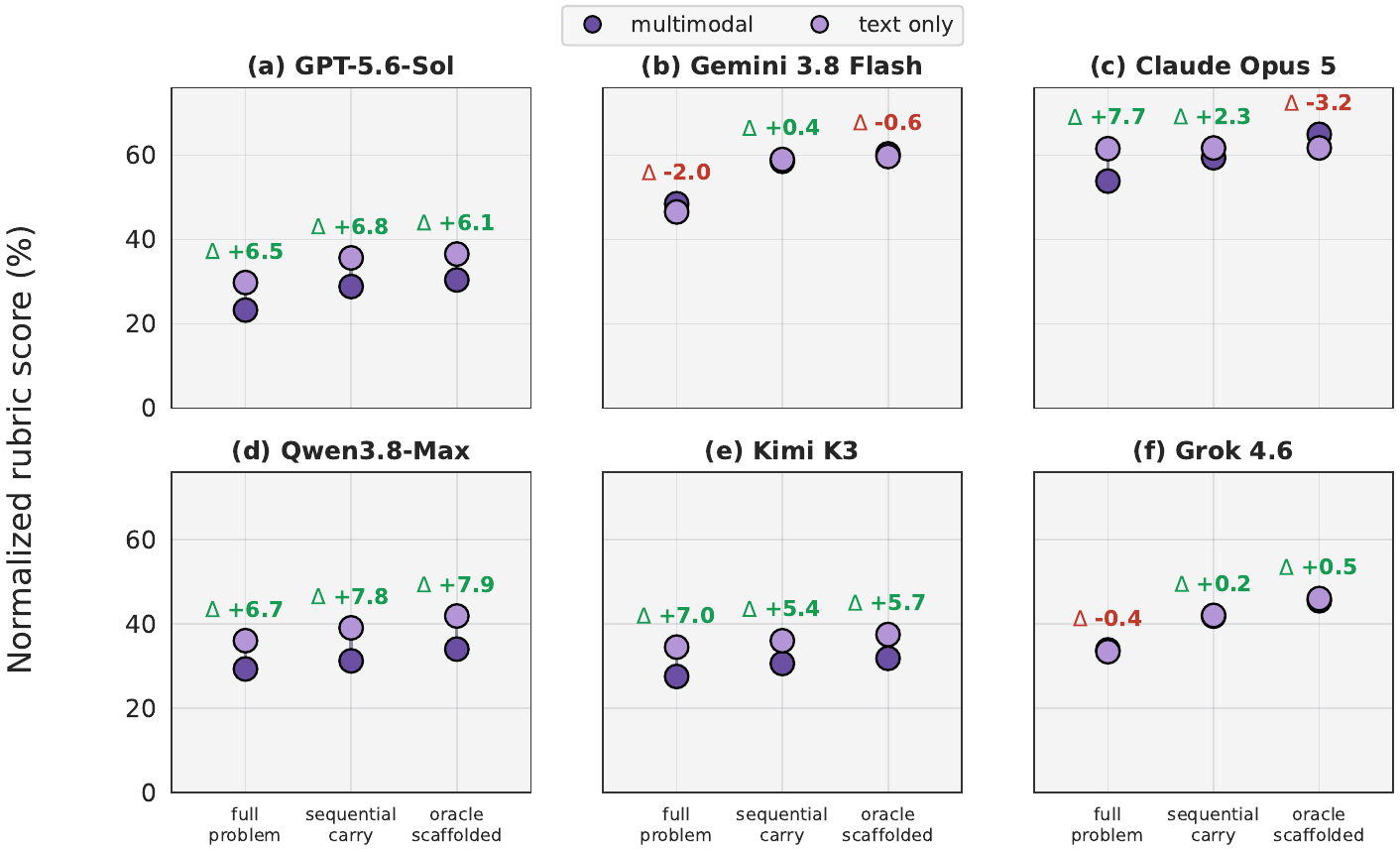}
\caption{Modality contrast for six of the seven models reported under both inputs, one panel per model, with $\Delta$ as text only minus multimodal in percentage points, positive values favoring text only.}
\label{fig:modality_contrast}
\end{figure}

The dumbbells in \Cref{fig:modality_contrast} carry the same contrast
visually, and the shape differs per
model. Qwen3.8-Max widens toward the step, 6.7 pp on the whole problem to
7.9 pp under oracle scaffolded, Kimi K3 peaks at the whole problem at 7.0
pp, dips to 5.4 pp at the carry, and settles at 5.7 pp, and GPT-5.6-Sol
peaks at the carry at 6.8 pp,
so the text advantage is a per-model profile rather than a constant tax.
Gemini 3.8 Flash concentrates its image advantage at the whole problem,
2.0 pp, gives back 0.4 pp at the carry, and regains 0.6 pp at the step,
so its advantage is an integration advantage, earned where the chain is
longest, step by step. Claude Opus 5 mirrors the profile, 7.7 pp for text
at the whole problem shrinking to 2.3 pp at the carry and flipping 3.2 pp
the other way at the step.

\subsection{Reader Effect}
\label{sec:exp:reader}

The reader reshapes the ranking model by model, and the direction is now
mostly down, nineteen of the twenty-seven metric cells negative. GLM-5.2
rises 28.5 pp on the whole problem and DeepSeek
V4 Pro rises 11.7 pp, but Claude Opus 5 falls 15.4 pp on the whole
problem, the deepest whole-problem fall in the roster, Gemini 3.8 Flash
falls 12.7 pp, Grok 4.6 falls
4.8 pp, Qwen3.8-Max falls 8.1 pp, and Kimi K3 falls 2.2 pp, so five of
the nine models score lower with the reader on the whole problem and
four score higher, as Table~\ref{tab:reader_effect} shows for every
metric, with GLM-5.2's 28.5 pp gain the largest move in either direction.

\begin{table}[t]
\centering
\small
\caption{Reader effect per model. Text + reader runs pair a text model with the Qwen3-VL image reader, which converts each image to text, and $\Delta$ marks text + reader minus text only, in percentage points, with positive values favoring the reader pairing.}
\label{tab:reader_effect}
\begin{tabularx}{\textwidth}{>{\raggedright\arraybackslash\hsize=1.8\hsize}X >{\centering\arraybackslash\hsize=0.91\hsize}X >{\centering\arraybackslash\hsize=0.91\hsize}X >{\centering\arraybackslash\hsize=0.91\hsize}X >{\centering\arraybackslash\hsize=0.91\hsize}X >{\centering\arraybackslash\hsize=0.91\hsize}X >{\centering\arraybackslash\hsize=0.91\hsize}X >{\centering\arraybackslash\hsize=0.91\hsize}X >{\centering\arraybackslash\hsize=0.91\hsize}X >{\centering\arraybackslash\hsize=0.91\hsize}X}
\toprule
\headrow Model & \multicolumn{3}{c}{\ra{\compA}} & \multicolumn{3}{c}{\rb{\compB}} & \multicolumn{3}{c}{\rc{\compC}} \\
\cmidrule(lr){2-4}\cmidrule(lr){5-7}\cmidrule(lr){8-10}
\headrow & text & reader & $\Delta$ & text & reader & $\Delta$ & text & reader & $\Delta$ \\
\midrule
DeepSeek V4 Pro & 25.0 & 36.7 & \textcolor{posgreen}{+11.7} & 25.0 & 36.8 & \textcolor{posgreen}{+11.8} & 27.6 & 37.0 & \textcolor{posgreen}{+9.4} \\
\bandrow GLM-5.2 & 23.5 & 52.0 & \textcolor{posgreen}{+28.5} & 24.9 & 52.2 & \textcolor{posgreen}{+27.3} & 25.0 & 52.4 & \textcolor{posgreen}{+27.3} \\
GPT-5.6-Sol & 29.8 & 31.1 & \textcolor{posgreen}{+1.3} & 35.6 & 33.6 & \textcolor{negred}{-2.1} & 36.5 & 34.0 & \textcolor{negred}{-2.6} \\
\bandrow Gemini 3.8 Flash & 46.5 & 33.8 & \textcolor{negred}{-12.7} & 59.0 & 45.1 & \textcolor{negred}{-13.9} & 59.7 & 49.7 & \textcolor{negred}{-9.9} \\
Claude Opus 5 & 61.5 & 46.2 & \textcolor{negred}{-15.4} & 61.7 & 53.9 & \textcolor{negred}{-7.8} & 61.7 & 54.7 & \textcolor{negred}{-7.0} \\
\bandrow Qwen3.8-Max & 36.0 & 27.9 & \textcolor{negred}{-8.1} & 39.0 & 29.2 & \textcolor{negred}{-9.8} & 41.9 & 31.5 & \textcolor{negred}{-10.4} \\
Kimi K3 & 34.5 & 32.3 & \textcolor{negred}{-2.2} & 36.0 & 32.3 & \textcolor{negred}{-3.7} & 37.5 & 32.4 & \textcolor{negred}{-5.0} \\
\bandrow Grok 4.6 & 33.4 & 28.6 & \textcolor{negred}{-4.8} & 42.0 & 29.0 & \textcolor{negred}{-13.0} & 46.0 & 31.1 & \textcolor{negred}{-14.9} \\
MiniMax M3 & 26.2 & 27.2 & \textcolor{posgreen}{+1.0} & 28.7 & 27.2 & \textcolor{negred}{-1.5} & 29.1 & 27.3 & \textcolor{negred}{-1.8} \\
\bottomrule
\end{tabularx}
\end{table}

Each delta in the table isolates the reader's contribution alone, because
it is measured against a fixed baseline, the same text run a model
produces without the reader. The pairing is the same Qwen3-VL module for
every model, which is what makes the deltas comparable across rows, and
the comparison answers a design question rather than a capability
question, whether an external vision module is a better way to deliver
images than the model's own eyes, and the roster answers it model by
model, with no single direction.

\begin{figure}[htb]
\centering
\includegraphics[width=\linewidth]{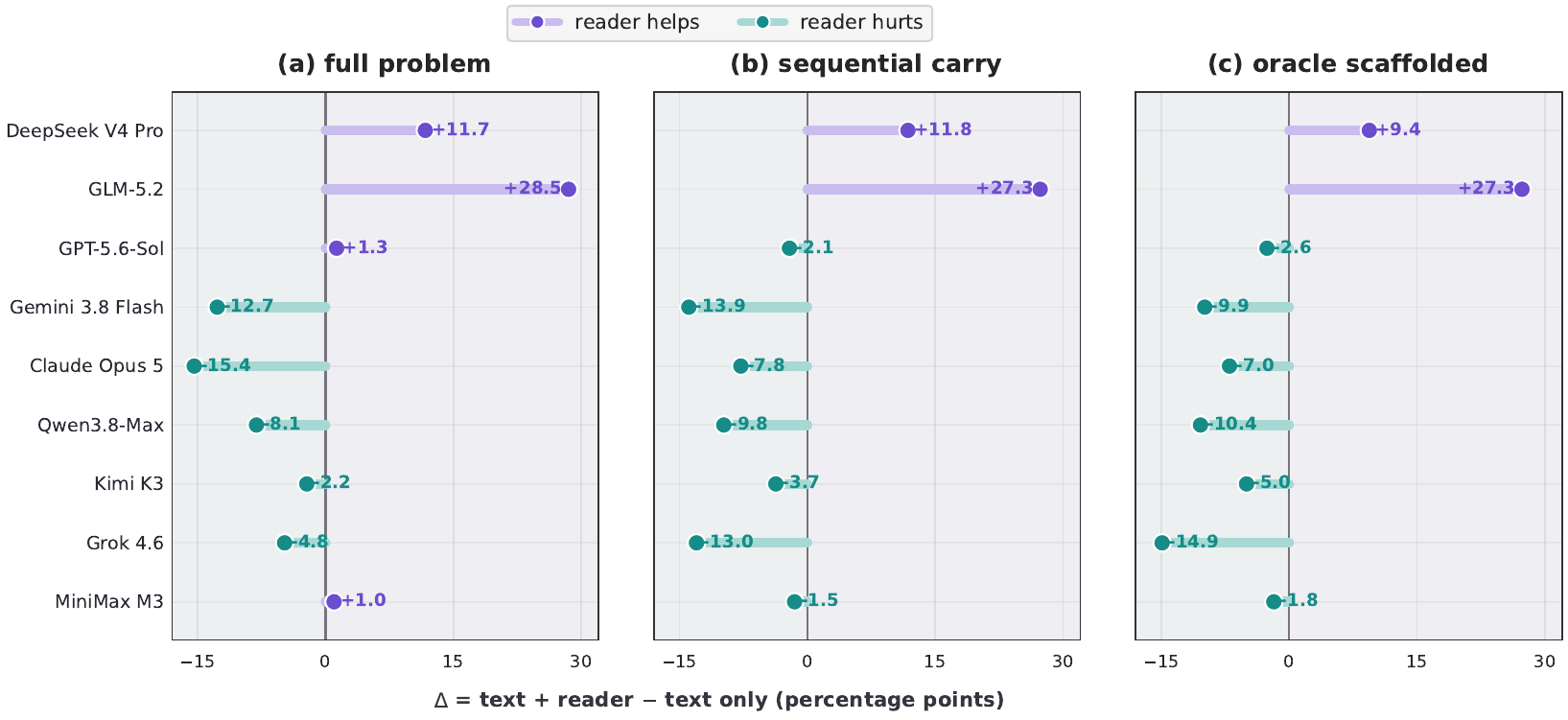}
\caption{Reader effect per model in three side-by-side metric panels. Deltas are text + reader minus text only in percentage points; purple marks reader gains and teal marks reader losses.}
\label{fig:reader_effect}
\end{figure}

The three side-by-side panels of \Cref{fig:reader_effect} make direction
and magnitude readable at a
glance. The marks extend right across all three metrics only for DeepSeek
V4 Pro and GLM-5.2, extend left for Gemini 3.8 Flash, Claude Opus 5,
Grok 4.6,
Qwen3.8-Max, and Kimi K3,
and switch direction between metrics for GPT-5.6-Sol and MiniMax M3, both
of which gain slightly on the whole problem and then lose a little on the
sequential metrics, the two spans crossing the zero line in the same
direction.

The reader also reshapes the ladder itself. For GLM-5.2, DeepSeek V4 Pro,
Kimi K3, and MiniMax M3 it flattens the ladder to under 0.6 pp, the
visual load absorbed before the model sees the problem, while for Gemini
3.8 Flash it sharpens the ladder to 15.9 pp, the steepest in the roster,
because the reader hands its strong text side a plain description and
strips away the images that made its multimodal run the best in the
roster, and for Claude Opus 5 it opens an 8.5 pp climb in a text line
that was flat at 0.2 pp. The deepest falls also come at the step, Grok
4.6 and Qwen3.8-Max
ending 14.9 and 10.4 pp down under oracle scaffolded, the two deepest
falls of the roster.

\subsection{Answers to the Two Questions}
\label{sec:exp:answers}

The two questions that \Cref{sec:intro} leaves open now carry answers.
Question I asks whether visual input adds understanding beyond text, and
the answer is now split. Three models gain from images on the whole
problem, Gemini 3.8 Flash by 2.0 pp, Grok 4.6 by 0.4 pp, and MiniMax M3
by 1.1 pp, while four lose by 6.5 to 7.7 pp, so the modality premium is
model-dependent, and the dividing line is the model's own vision.
Question II asks whether failure originates in global strategy or local
execution, and the ladder answers uniformly, the trough sits at the whole
problem in every one of the twenty-five reported runs, so the strategy
level is where every model loses the most, though its depth varies from
0.1 pp in the flattened reader runs to 13.2 pp for Gemini 3.8 Flash from
text alone. The reader effect, reported in
Table~\ref{tab:reader_effect}, adds a third finding neither question
anticipated, it is model-dependent, and it moves the ranking by 28.5 pp
up for GLM-5.2 on full problem against 15.4 pp down for Claude Opus 5 on
the whole problem, a 43.8 pp swing, the largest in the roster.

\section{Conclusion}
\label{sec:discussion}


Chemistry evaluation can no longer certify only \emph{that} a model's final
answer matches; it must certify \emph{where} a solution chain goes wrong,
and \bench\ is the instrument that makes that certification mechanical.
Its core contribution is the three protocols sharing one scoring unit,
\ra{\compA} for whole-problem integration, \rb{\compB} for sequential error
propagation, and \rc{\compC} for local reasoning, graded on rubric points
with partial credit, deterministic chemistry-equivalence checks for formal
answers, an audited independent judge for flexible ones, and image
visibility isolated so that the model never reads its own grading key.
Evaluating nine frontier models across three input modes, with
twenty-five completed runs, yields three stable findings. Whole-problem
integration is the largest measured cost, with the unassisted whole
problem scoring below both sequential metrics in every reported run,
23.2 vs.\ 28.8 and 30.4 for GPT-5.6-Sol with images and 46.5 vs.\ 59.0
and 59.7 for Gemini 3.8 Flash with text; modality is priced rather than
assumed, and the roster splits in two, Gemini 3.8 Flash, Grok 4.6, and
MiniMax M3 each gain from real images on the whole problem, 2.0, 0.4,
and 1.1 pp, while GPT-5.6-Sol, Qwen3.8-Max, Kimi K3, and Claude Opus 5
each lose 6.5 to
7.7 pp by reading them; and the reader effect is model-dependent, lifting
GLM-5.2 by 28.5 pp on the whole problem while lowering Claude Opus 5 by
up to
15.4 pp, a 43.8 pp swing across the roster. The best reported score, 64.9
for Claude Opus 5 under oracle scaffolded with images, still leaves
more than a third of the rubric credit unclaimed. What a chemistry
benchmark must certify, then, is exactly what this one measures, not a
number, but a diagnosis, one that tells the next model exactly which link
of the chain to strengthen, and that is the direction in which
chemistry-capable language models can now be improved.

{\small
\bibliography{references}
\bibliographystyle{unsrtnat}}

\newpage
\appendix

\end{document}